\documentclass[10pt,twocolumn,letterpaper]{article}

\usepackage{iccv}
\usepackage{times}
\usepackage{epsfig}
\usepackage{graphicx}
\usepackage{amsmath}
\usepackage{amssymb}
\usepackage{xcolor}
\usepackage{soul}
\usepackage{url}

\usepackage[utf8]{inputenc}
\usepackage{booktabs}
\usepackage{verbatim}
\usepackage{algorithmic}
\usepackage{algorithm}
\usepackage{bm}

\iccvfinalcopy 

\def\iccvPaperID{7} 

\ificcvfinal\fi

\begin{document}
	
	\title{Structuring Autoencoders}
	
	\author{Marco Rudolph \hspace{2cm} Bastian Wandt \hspace{2cm} Bodo Rosenhahn\\
		Leibniz Universit\"at Hannover\\
		{\tt\small {\{rudolph, wandt, rosenhahn\}@tnt.uni-hannover.de}}
	}
	
	\maketitle
	
	\begin{abstract}
		
		In this paper we propose \emph{Structuring AutoEncoders (SAE)}.
		SAEs are neural networks which learn a low dimensional representation of data and are additionally enriched with a desired structure in this low dimensional space.
		While traditional Autoencoders have proven to structure data naturally they fail to discover semantic structure that is hard to recognize in the raw data. 
		The SAE solves the problem by enhancing a traditional Autoencoder using weak supervision to form a structured latent space.
		
		
		
		In the experiments we demonstrate, that the structured latent space allows for a much more efficient data representation for further tasks such as classification for sparsely labeled data, an efficient choice of data to label, and morphing between classes. 
		To demonstrate the general applicability of our method, we show experiments on the benchmark image datasets MNIST, Fashion-MNIST, DeepFashion2 and on a dataset of 3D human shapes. 
	\end{abstract}
	
	\section{Introduction and Related Work}
	Data structuring is widely used to analyze, visualize and interpret information. 
	A common approach is to employ autoencoders \cite{hinton2006reducing} which try to solve this task by structuring data in an unsupervised fashion.
	Unfortunately, they tend to focus on the most dominant structures in the data which not necessarily incorporate meaningful semantics.
	In this paper we propose Structuring AutoEncoders (SAE) which enhance traditional autoencoders with weak supervision.
	These SAEs can enforce a structure in the latent space desired by a user and are able to separate the data according to even subtle differences. 
	The structured latent space opens up a variety of applications: 
	\begin{figure}[h]
		\centering
		\includegraphics[width=0.45\textwidth]{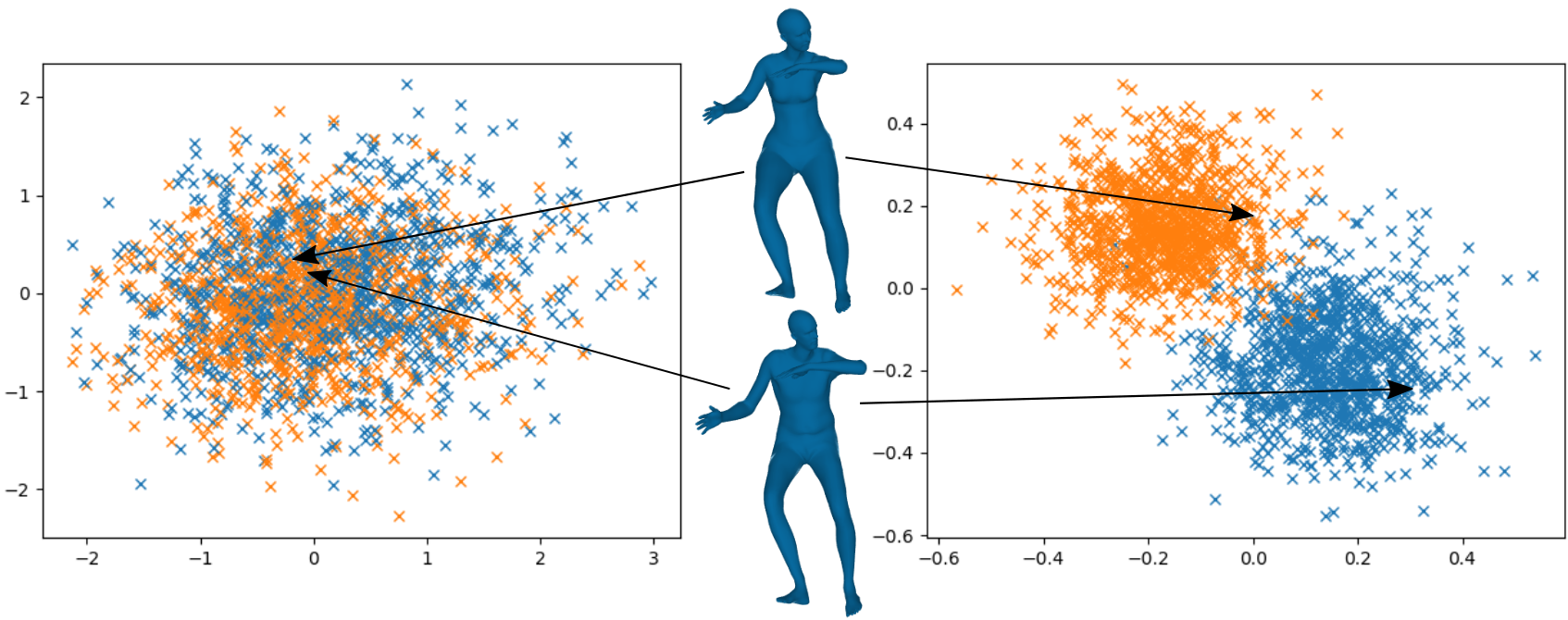} 
		\caption{Latent spaces of the autoencoders for the 3D HumanPose database. The colors are given by the gender, male and female. Left: Confused latent space when using a traditional autoencoder. Right: Clustered structure in latent space when using the SAE.
		}
		\label{fig:StrAe}
	\end{figure}
	\begin{enumerate}
		\item Improving classification accuracy on datasets where only a small number of data points is labeled.
		\item Finding the most important unlabeled data points for giving labeling recommendations.
		\item An interpretable latent space for data visualization.
		\item Morphing between properties that are hidden in the data.
	\end{enumerate}
	
	
	The focus of this work is to transfer data into an organized structure that reflects a meaningful representation. 
	To achieve this, it is necessary to uncover even subtle semantic characteristics of data. 
	As an enhancement of linear factorization models \cite{grasshof2017apathy}, the idea of autoencoders as a tool to naturally uncover structures has been part of research on neural networks for decades \cite{lecun-89,Bourlard1988,Vincent2010StackedAE}.
	They are commonly used to learn representative data codings and usually consist of a neural network having an encoder and a decoder.
	The encoder maps the data points through one or more hidden layers to a low dimensional latent space from where the decoder reconstructs the input.
	However, this representation is not necessarily meaningful in terms of the underlying semantics and cannot discover well hidden structures.
	There are other variants of Autoencoders which enforce a specific distribution in the latent space, either by a variational approach \cite{Kingma13} or by applying a discriminator network on the latent space known as Adversarial Autoencoders \cite{Makhzani16}. 
	Other works focussed on getting disentangled representations of data in the latent space \cite{Kulkarni15,donahue18, gu2019mask,awiszus2019faces}.
	There are several other variants that find additional constraints on the latent variables, mostly for specific applications \cite{roberts2018hierarchical,chen2018subspace,Song13,Liu16,Carr15,Li18,Chen12}.
	However, analysis of hidden structures is rarely considered. 
	Our approach solves this task by improving traditional autoencoders with a weak supervision using only a very small amount of additionally labeled data which represents the desired formerly well-hidden semantics.
	Furthermore, we propose a method to extend this small set of labels efficiently by determining critical examples that are most meaningful to improve classification.
	Comparing common classification networks to our approach, they can be interpreted as the omission of the decoder network.
	
\begin{figure}
		\centering
		\includegraphics[width=0.45\textwidth]{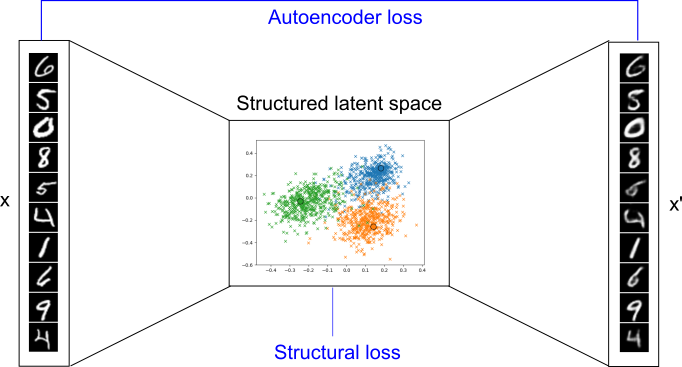} 
		\caption{Our Structuring AutoEncoder (SAE) projects data into a structured latent space. It uses Multidimensional Scaling to calculate the class centers in the latent space. By applying an additional structural loss the SAE maintains distances between the classes according to a desired metric. Losses are colored in blue.}
		\label{fig:diagram}
	\end{figure}

	
	As an example we consider the separation of male and female 3D body shapes which are in different poses.
	The obvious structure in the data is the pose of the body shapes since the variation in pose is a lot stronger compared to variation in the gender regarding the reconstruction error.
	In fact, passing the data through a traditional autoencoder it will mix male and female data points as can be seen on the left hand side of Fig.~\ref{fig:StrAe}.
	To assist the autoencoder to separate the data points into male and female we define distances between different classes.
	These distances shall be maintained in the latent space while training the SAE.
	Following the example we specify a distance of $1$ between the male and female class.
	The distance metric is freely customizable to a desired task.
	The right image of Fig.~\ref{fig:StrAe} shows a much better organized latent space obtained by the SAE. 
	Interestingly, there is only a marginal increase of the reconstruction error when using the SAE compared to standard autoencoders.
	For ordering data with respect to the relative distance measures in this work Multidimensional Scaling (MDS) is applied \cite{MDS1}.
	Alternative approaches such as t-SNE, which is based on a Stochastic Neighbor Embedding \cite{tSNE1,tSNE2} or Uniform Manifold Approximation and Projection (UMAP) \cite{umap} are conceivable.
	These methods can be used to visualize the level of similarity of individual examples of a dataset and can be seen as related ordination techniques which is used in information visualization.
	To preserve desired distances in the latent space we use MDS in this work.
	By applying MDS on sparsely known labels of the training set, it allows to structure the data in such a fashion, that data points with the same labels have a small distance in the latent space, whereas data belonging to different labels are enforced to keep a certain distance.
	This is formulated as the structural loss in addition to the decoders reconstruction error.
	A diagram of the proposed autoencoder training including a structured latent space visualization and the used losses is shown in Fig.~\ref{fig:diagram}.
	
	We show experiments on the benchmark dataset MNIST \cite{lecun2010} which we randomly decompose into three classes.
	The results underline the fact that the SAE efficiently separates the latent space according to a freely selected structure that is invisible the raw data.
	Moreover, using only a very sparse set of data (6000 labeled samples) the SAE outperforms comparable neural networks trained solely for the classification task.
	These results are confirmed on the recent more diverse dataset Fashion-MNIST \cite{fmnist} and our own dataset of 3D meshes of human body shapes.
	A real-world application is shown on the recently published DeepFashion2 \cite{deepfashion2} dataset where our SAE outperforms comparable classifiers.
	Additionally, we show that our guided labeling approach only needs $600$ training samples combined with the $100$ most meaningful samples that are automatically detected to achieve good classification results.
	This provides a tool to significantly reduce labeling time and cost.

	Summarizing, our contributions are:
	\begin{itemize}
		\item An autoencoder that structures data according to given classes and preserves distances present in the label space.
		\item A method to deal with sparsely labeled data while preventing the overfitting of traditional approaches.
		\item Better classification performance than comparable neural networks trained for classification using the same amount of training data.
		\item Similar training performance (reconstruction loss) with and without structured training.
		\item A technique to improve the labeling efficiency by determining critical data points.
	\end{itemize}
	
	\section{Structuring Autoencoder}
	We assume that the input data can be separated into several classes which are not obvious in the data itself.
	These classes are only known for a small fraction of the input data.
	We further assume that the data can be projected to a latent space that preserves the distances between the classes.
	As a toy example we separate the Fashion-MNIST dataset \cite{fmnist} into the three classes \textit{summer clothes} (top, sandals, dress and shirt), \textit{winter clothes} (pullover, coat and ankle boot), and \textit{all-year fashion} (sneaker, trousers, bag).
	The left hand side of Fig.~\ref{fig:compare_decompositions} shows the latent space of this.
	Here, as an example we define an equal distance between the classes.
	Obviously, the season depending decomposition is not given by the data itself.
	The following sections describe the proposed autoencoder architecture and training.
	Algorithm~\ref{alg:ae} describes the steps for training the network.
	
	\begin{algorithm}
		\caption{Autoencoder training}
		\label{alg:ae}
		\begin{algorithmic}
			\STATE $\bm{X} \gets$ training samples
			\STATE $\bm{D} \gets$ distances
			
			\WHILE{no convergence}
			
			\STATE $\bm{Z}=\bm{f}_{enc}(\bm{X})$ \COMMENT{project $\bm{X}$ into latent space}
			\STATE $\bm{Z}^*= MDS(\bm{Z})$ \COMMENT{calculate desired positions}
			\STATE $\bm{ZZ}^+=\bm{USV}^T$ \COMMENT{singular value decomposition}
			\STATE set all singular values $\geq 0$ to $1$
			\STATE $\bm{R}=\bm{US}^*\bm{V}^T$ \COMMENT{calculate ideal rotation}
			\STATE $\tilde{\bm{Z}}=\bm{R}\bm{Z}^*$ \COMMENT{final positions in latent space}
			
			\STATE train SAE with loss $\mathcal{L}_{SAE}(\bm{x}, \tilde{\bm{z}})$ and $\mathcal{L}_{AE}(\bm{x}, \bm{f}_{ae}(\bm{x}))$
			\ENDWHILE
		\end{algorithmic}
	\end{algorithm}

	\subsection{Architecture and Loss Functions}
	Our method is not restricted to a specific autoencoder architecture.
	That means every architecture can be applied, for instance fully connected, (fully) convolutional, or adversarial autoencoders.
	We define two loss functions.
	The first loss 
	\begin{equation}
	\mathcal{L}_{AE}(\bm{x}, \bm{f}_{ae}(\bm{x})) = \| \bm{x} - \bm{f}_{ae}(\bm{x})\|_2^2
	,
	\end{equation}
	is the mean squared error (MSE) between the input $\bm{x}$ and the output of the autoencoder $\bm{f}_{ae}(\bm{x})$ as it is commonly used.
	With $\bm{f}_{enc}(\bm{x})$ as the function of the encoder that projects $\bm{x}$ to the latent space a structural loss is defined as 
	\begin{equation}
	\mathcal{L}_{S}(\bm{f}_{enc}(\bm{x}), \tilde{\bm{z}}) =  \|\bm{f}_{enc}(\bm{x}) - \tilde{\bm{z}}\|_2^2
	.
	\end{equation}
	It is calculated by the MSE between the latent values $\bm{f}_{enc}(\bm{x})$ and the desired locations $\tilde{\bm{z}}$ in the latent space that are calculated at each iteration.
	The estimation of these locations using Multidimensional Scaling is described later in Sec.~\textbf{\ref{sec:trainingloop}}. 
	This gives the combined loss
	\begin{multline}
	\label{eqn:loss}
	\mathcal{L}_{SAE}(\bm{x}, \tilde{\bm{z}}) = \\
	\gamma \mathcal{L}_{S}(\bm{f}_{enc}(\bm{x}), \tilde{\bm{z}}) + 
	(1-\gamma)\mathcal{L}_{AE}(\bm{x}, \bm{f}_{ae}(\bm{x}))
	,
	\end{multline}
	with $\gamma =[0,1]$ as the balancing parameter between the two losses.
	Note that $\gamma=0$ corresponds to the traditional autoencoder training while a higher value of $\gamma$ gives a higher importance to the structural loss.
	In section \ref{effect_mds} the influence of $\gamma$ is analyzed and its choice for experiments is explained.
	For unlabeled data $\mathcal{L}_{SAE} = \mathcal{L}_{AE}$ is considered since there is no $\tilde{\bm{z}}$ defined.
	
	\subsection{Initialization}
	\label{sec:init}
	Following the toy example from above a distance matrix $\bm{D}$ between the three classes is calculated where each row and column marks a training sample and the entries are the distances.
	Here, we can define an equal distance (e.g. of $1$) between different classes.
	The intra class distance is $0$.
	Since the distances between the classes stay the same during training the distance matrix only needs to be calculated once.
	
	\subsection{Structuring the latent space}
	\label{sec:trainingloop}
	The autoencoder is trained iteratively.
	In every iteration the data $\bm{x}$ is projected into the latent space by the encoder which gives the latent variables 
	
	\begin{figure*}
		\centering
		\includegraphics[width=0.85\textwidth]{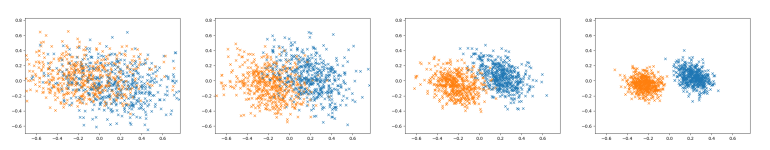} 
		\caption{Visualization of the iteration steps. 
			With each iteration the two classes are separated better in the latent space. 
			The images show the same two dimensions in every step for the 3D body shape dataset.}
		\label{fig:convergence_lat}
	\end{figure*}
	
	\begin{equation}
	\bm{z} = \bm{f}_{enc}(\bm{x})
	.
	\end{equation}
	This is done for the complete training set. 
	By stacking all $\bm{z}$ vectors we obtain the matrix $\bm{Z}$.
	To calculate the desired latent positions $\tilde{\bm{Z}}$ we apply \textit{Multidimensional Scaling} (MDS) \cite{Kruskal1964} to the distance matrix $\bm{D}$ that is defined in Section~\ref{sec:init}. 
	MDS is able to arrange data points in a space of an arbitrary dimension in a way that the given distances should be preserved.
	The \textit{Shepard-Kruskal algorithm} \cite{Kruskal1964} is an iterative method to find such an arrangement.
	After an initialization the stress between the actual and the given distance measures is minimized until a local minimum is found.
	In contrast to manually setting the desired latent locations the MDS can automatically adapt to the data and therefore to the training process.
	This results in a target matrix of locations $\bm{Z}^*$ in the latent space. 
	
	Since there is an infinite number of possible target locations and we want to compute locations close to $\bm{Z}$ the MDS algorithm is initialized with them. 
	To get the best possible target locations an orthogonal alignment \cite{Schonemann1966} is applied to $\bm{Z}^*$ to best fit $\bm{Z}$.
	Naturally, MDS results in centralized data points. 
	Therefore, we only need to compute the ideal rotation around the origin.
	Let $\bm{P}$ be a projection matrix that projects $\bm{Z}^*$ to $\bm{Z}$ by 
	\begin{equation}
	\bm{Z}=\bm{P}\bm{Z}^*
	.
	\end{equation}
	We assume that there is a Moore-Penrose-Inverse $\bm{Z}^+$ of $\bm{Z}^*$ with $\bm{Z}^*\bm{Z}^+=\bm{I}$, where $\bm{I}$ is the identity matrix. 
	This states true if there are more data points than latent dimensions, which is always the case in a meaningful experimental setting.
	The singular value decomposition of $\bm{P}^* = \bm{Z}\bm{Z}^+$ gives
	\begin{equation}
	\bm{P}^*=\bm{USV}^T
	.
	\end{equation}
	A new matrix $\bm{S}^*$ is defined by copying $\bm{S}$ and setting all nonzero singular values to $1$.
	Then the ideal rotation $\bm{R}$ can be found by 
	\begin{equation}
	\bm{R}=\bm{US}^*\bm{V}^T
	.
	\end{equation}
	The desired latent positions are calculated by
	\begin{equation}
	\tilde{\bm{Z}}=\bm{R}\bm{Z}^*
	.
	\end{equation}
	With these target locations the autoencoder is trained batch-wise for a complete epoch. 
	After the epoch the steps in this section are repeated until convergence.
	The data in the latent space during the training steps is visualized in Fig.~\ref{fig:convergence_lat}.
	
	
	
	
	
	\section{Experiments}
	\begin{figure}[tp]
		\centering
		\includegraphics[width=0.48\textwidth]{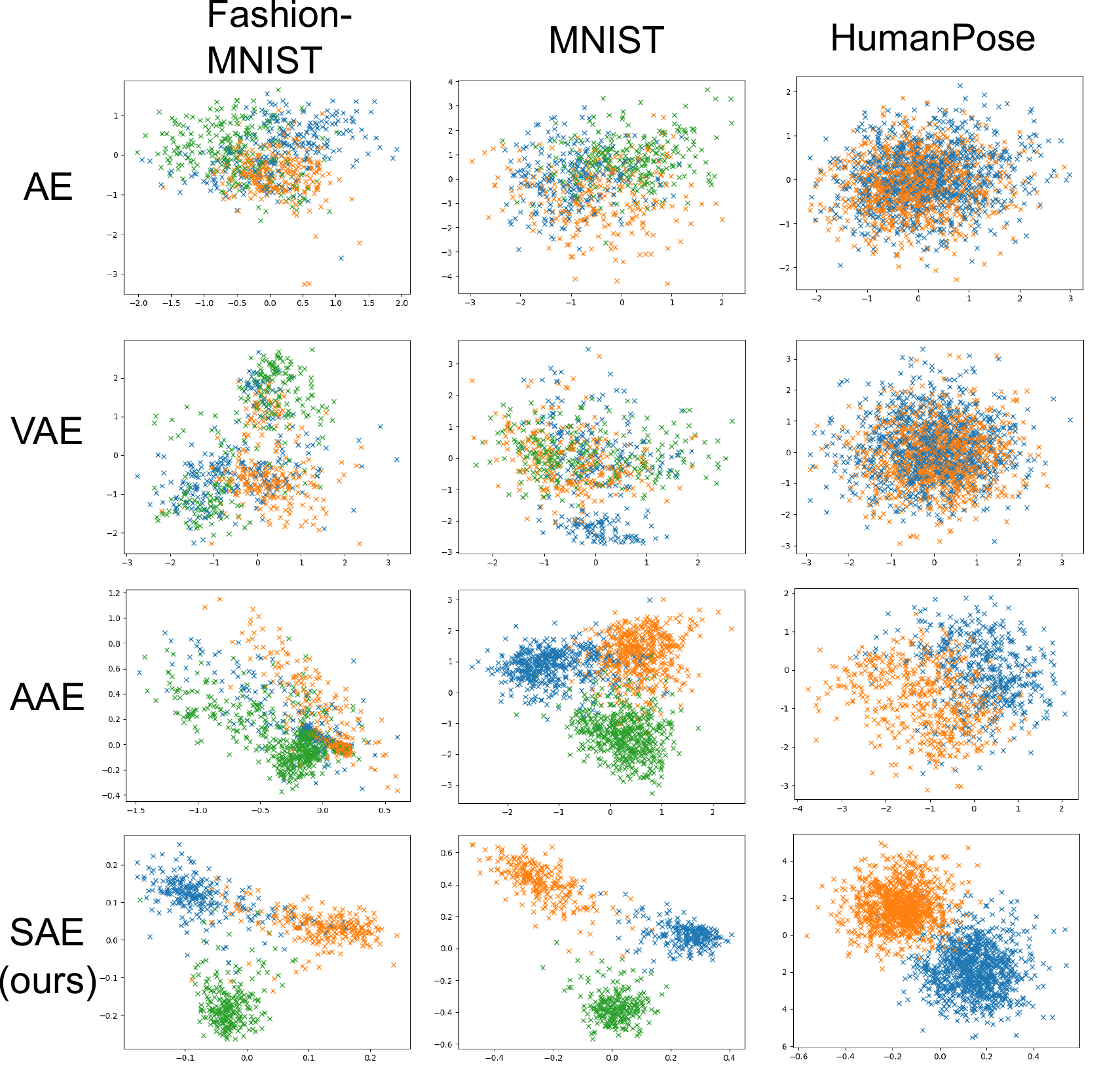} 
		\caption{The scatterplots show 2D projections of the latent space when using different types of autoencoders. For each instance an appropriate projection was chosen. Points of the same color represent samples from the same class. It can be clearly seen that only a Structuring Autoencoder is able to separate the latent variables well.}
		\label{fig:3x3projections}
	\end{figure}
	
	\begin{figure}[tp]
		\centering
		\includegraphics[width=0.49\textwidth]{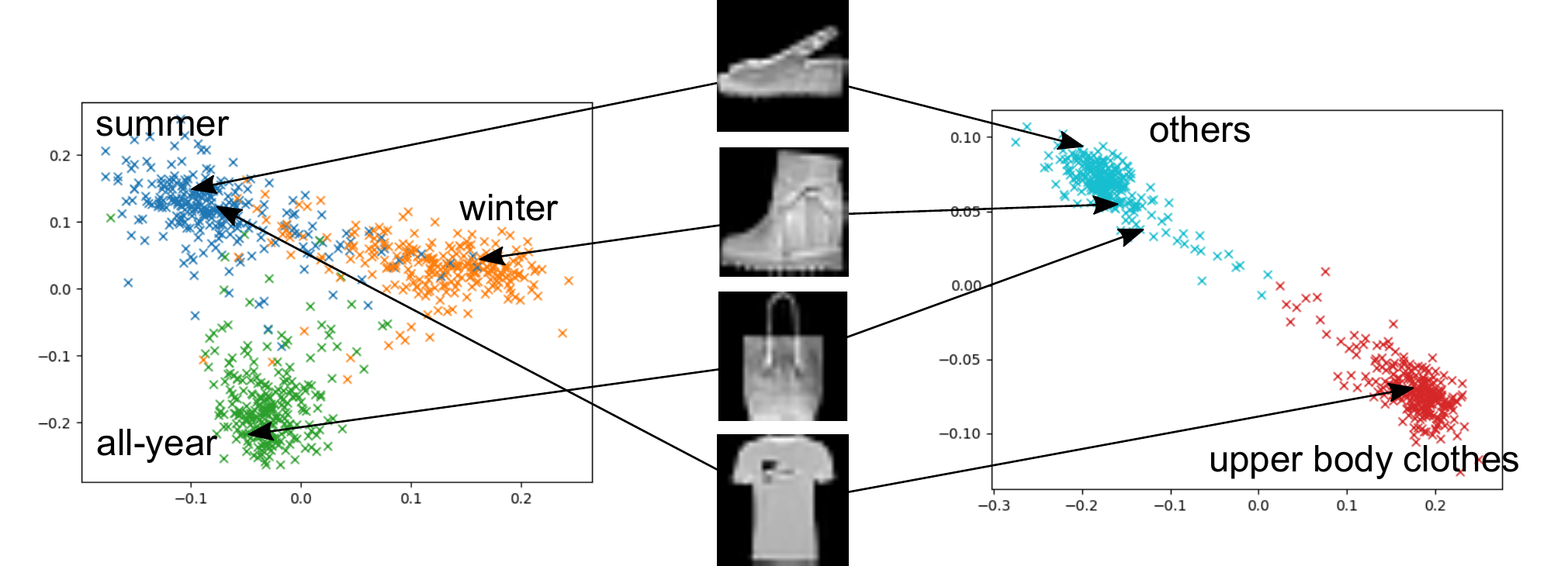} 
		\caption{Comparison of two projections of the latent space using different decompositions of the data. Note that the distance of two samples is highly influenced by the chosen decomposition so this setting is a method to individually control data.}
		\label{fig:compare_decompositions}
	\end{figure}

	We show the performance of our algorithm in several experiments using diverse datasets including images and vector data.
	The evaluation is done on the benchmark datasets MNIST \cite{lecun2010}, the recently published fashion datasets Fashion-MNIST \cite{fmnist} and DeepFashion2 \cite{deepfashion2}, and our own 3D body shape dataset created using SMPL \cite{smpl2015}.
	It is important to note, that we focus on artificially set classes.
	That means we try to find clusters that are not evident or barely visible in the original data, e.g. a season depending decomposition of Fashion-MNIST.
	Furthermore, we show that the SAE generalizes very well if only a small subset of the training data is used.
	Since we achieve a clear separation of the defined classes in the latent space after training we can fit an optimal hyperplane between the classes using Support Vector Machines \cite{vapnik74theory}.
	This allows for the definition of a classification error considering the separation in the latent space.
	We further use the term \textit{reconstruction error} as the root-mean-square error (RMSE) between the input and output of the autoencoder.
	We only train on unaugmented data in all our experiments.
	This allows for a fair performance comparison between different classifiers even for data where no augmentation is possible, e.g. the 3D body shape data.
	We are aware of the fact, that state-of-the-art classification performance cannot be completely reached without data augmentation.
	However, we want to emphasize that the focus of the paper is on semantically structuring the latent space of autoencoders and not on state-of-the-art classification results on benchmark datasets.
	Therefore, we use standard fully connected and convolutional neural networks for all experiments and compare against comparable classification networks.
	This means the classification network uses the same architecture as the encoder of the SAE to be compared plus a fully connected output layer.
	
	\begin{figure*}
		\includegraphics[width=0.95\textwidth]{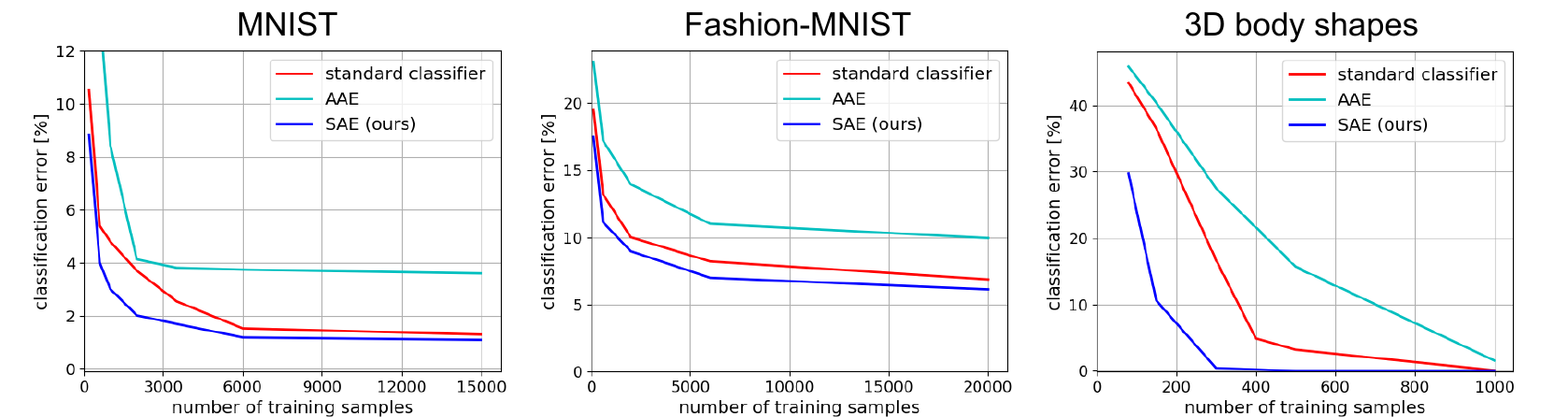} 
		\caption{Test error for different sizes of the training set without using data augmentation. The SAE outperforms a comparable traditional neural network and an adversarial autoencoder on each of the datasets significantly, especially if the number of labeled training samples is low.}
		\label{fig:samples_plot}
	\end{figure*}
	
	\subsection{Datasets and Neural Networks}
	To show an example on a well-known benchmark dataset we randomly divide MNIST into three classes $A=(0,1,9)$, $B=(4,6,8)$, and $C=(2,3,5,7)$.
	As a more realistic example we evaluate on the Fashion-MNIST dataset which was published in 2017 to have a benchmark which is a lot harder than the old original MNIST.
	It consists of a training set of 60,000 examples and a test set of 10,000 examples of various fashion items divided into 10 classes.
	According to the authors these images reflect real world challenges in computer vision better than the original MNIST dataset.
	We split Fashion-MNIST into the three classes \textit{summer clothes} (top, sandals, dress and shirt), \textit{winter clothes} (pullover, coat and ankle boot), and \textit{all-year fashion} (sneaker, trousers, bag).
	
	For both datasets, MNIST and Fashion-MNIST, a convolutional neural network is used for the encoder. It consists of three $3\times 3$ convolutional layers ($8$, $16$, $32$ filters), ReLU activation and pooling layers.
	The latent space has a dimension of $10$ for MNIST and $64$ for Fashion-MNIST.

	We used a subset of DeepFashion2 dataset where we only considered skirts and shorts to show the behaviour in borderline cases. For the encoder we use the convolutional part of the original VGG implementation \cite{SimonyanZ14a} and a latent space size of $192$.
	In all networks the decoder always mirrors the encoder.
	
	To show general applicability for different types of data we create a 3D HumanPose dataset that consists of randomly created human models with $3000$ male an $3000$ female meshes in various poses and body shapes using SMPL \cite{smpl2015}. 
	We only use the $x,y,z$ coordinates of the $6890$ vertices for training by stacking them in a vector.
	Since the data points are in vectorial form we use a fully connected network consisting of two dense layers $2048$ and $256$ neurons, respectively.
	The latent space has $30$ dimensions.
	This covers a variety of data from simple images (MNIST) and more complicated image (Fashion-MNIST) to data in vectorial form (3D HumanPose) and different network architectures.
	Note that our approach is flexible such that an arbitrary network structure can be applied for the encoder and decoder networks.
	

	\subsection{Structure Analysis}
	As already mentioned, some structure cannot be detected by traditional autoencoders because it is hidden in the data.
	This effect can be visualized easily by projecting into the latent space.
	Fig. \ref{fig:3x3projections} compares 2D projections of standard autoencoders (AE), variational autoencoders (VAE) and our proposed Structuring Autoencoders (SAE) for all datasets.
	Standard autoencoders barely show any structure in the form of clusters, whereas a slight clustering of samples of the same class can be observed when using variational autoencoders.
	However, the desired clear separation cannot be seen at all while our SAE provides a clean structured latent space.
	These examples use a fixed distance of $1$ between classes.
	However, the inter class distance can be freely defined.
	Additionally, also the decomposition of the data is free of choice.
	For example Fashion-MNIST can be decomposed in another way, e. g. differentiating between clothes worn at the upper body and other fashion items.
	Fig.~\ref{fig:compare_decompositions} compares the projections of the resulting latent space using this decomposition alongside the previously used one (summer, winter, all-year).

	\subsection{Improved Classification}
	Since the autoencoder separates the data in the latent space it is possible to train a simple linear classifier on the latent space.
	We show that a linear SVM trained on the latent variables achieves a better accuracy compared to a neural network of similar structure as the encoder.
	Since the SAE enforces a latent space that can be decoded overfitting is prevented even if only a small amount of training data is used.
	Fig.~\ref{fig:samples_plot} shows the error on the test set with different numbers of labeled samples compared to an adversarial autoencoder and a neural network solely trained for classification.
	For the training of the adversarial autoencoder we performed the semi-supervised method described in Section 2.3 of the corresponding paper \cite{Makhzani16} and applied SVM after training.
	It can be clearly seen that the SAE outperforms traditional classification networks on MNIST, Fashion-MNIST, and 3D HumanPose, especially when using only a few samples.
	
	Note that all experiments are done without data augmentation.
	For comparison, when applying data augmentation to the training data we achieve classification rates of 99.04\% on MNIST using only $6000$ samples.
	
	\begin{figure}[tp]
		\centering
		\includegraphics[width=0.45\textwidth]{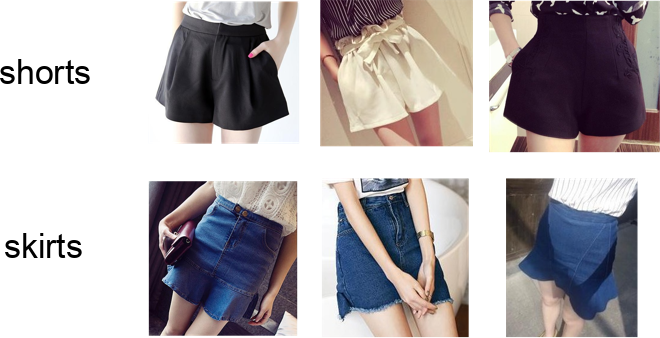} 
		\caption{Examples from the DeepFashion2 dataset where the class membership is visually hard to determine or features of the opposite class occur. In contrast to traditional classifiers the SAE assigns meaningful low confidence values to these samples.}
		\label{fig:critical_examples}
	\end{figure}
	
	\begin{figure}[tp]
		\centering
		\includegraphics[width=0.45\textwidth]{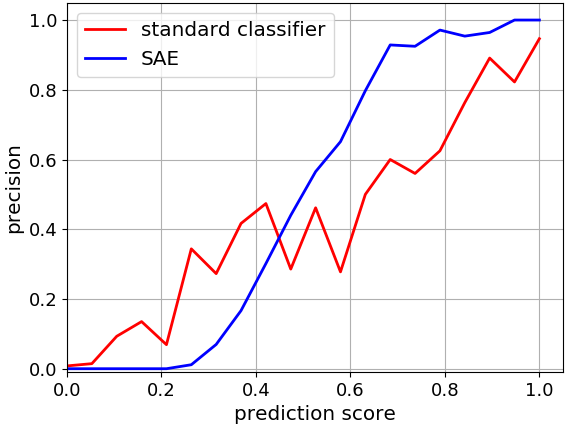} 
		\caption{Relation between the prediction score and the actual precision which is computed over samples from binned sets of prediction scores. 
			Contrary to the noisy plot of the standard classifier, the smooth SAE plot shows that there is a clear mapping between prediction scores and the actual precision.
			Thus it is evident that the scores provided by our SAE are much more reliable for critical decisions.}
		\label{fig:confidence_comparison}
	\end{figure}

	\subsection{Decision Confidence}
	Traditional neural networks used for classification aim to predict a class with high confidence mostly applying a softmax activation in the last layer.
	As a result their decision confidences appear to be relatively high even if the actual decision is uncertain.
	Our SAE avoids the uncertain predictions and gives a meaningful and interpretable confidence measurement.
	In real-world applications, for instance reflected by the DeepFashion2 \cite{deepfashion2} data set, there are several samples that are hard to assign to one class because of occlusions or the presence of features from several classes.
	Therefore, it is desirable to have expressive prediction scores.

	\begin{figure}[tp]
		\centering
		\includegraphics[width=0.48\textwidth]{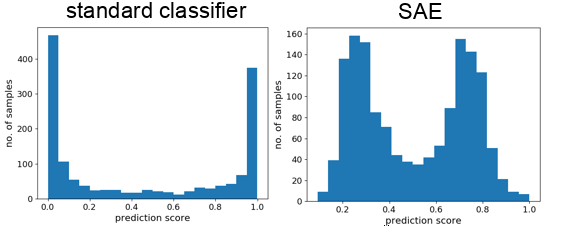} 
		\caption{Histogram of prediction scores when using a standard classifier and our SAE. While the standard classifier tends to predict scores near $0$ and $1$, the SAE outputs are more uniformly distributed over the interval to reflect the confidence better.}
		\label{fig:score_frequency}
	\end{figure}
	
	\begin{figure}[tp]
		\centering
		\includegraphics[width=0.4\textwidth]{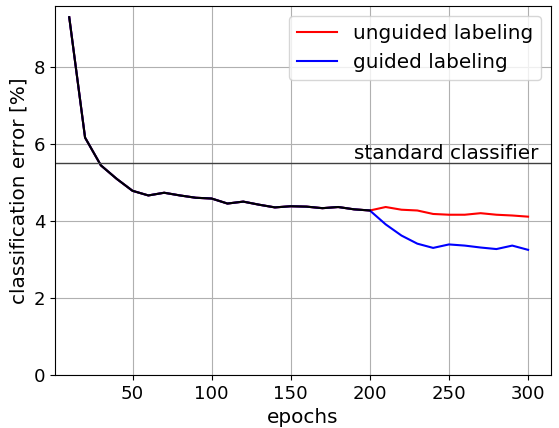} 
		\caption{Comparison of the guided and unguided sampling approach for the MNIST initially trained on $600$ samples. In epoch $200$ the $100$ most uncertain assigned data points according to the SAE were added. The \textit{standard classifier} threshold is a CNN of comparable structure as the encoder network which is solely trained for classification.}
		\label{fig:guided_sampling}
		\vspace{-10pt}
	\end{figure}
	
	\begin{figure*}[tph!]
		\centering
		\includegraphics[width=0.95\textwidth]{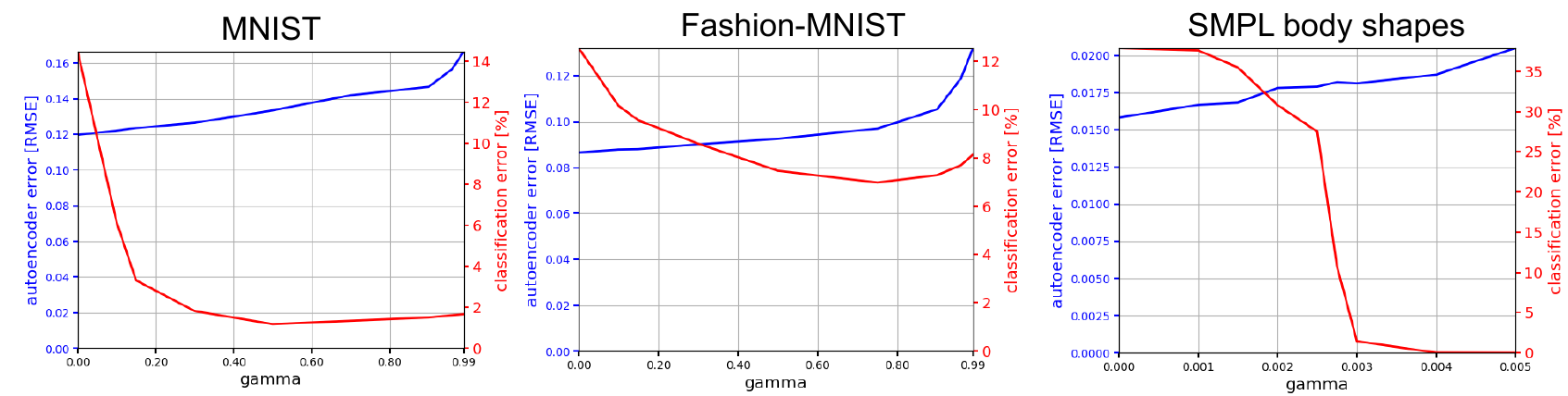} 
		\caption{Influence of the balancing parameter $\gamma$ on the autoencoder error and the classification error. For MNIST and Fashion-MNIST only $6000$ labeled training samples (10\% of the data) were used. The training set of 3D body shape dataset consists of $1000$ body shapes.}
		\label{fig:gamma_plot}
		\vspace*{-5pt}
	\end{figure*}
	
	For example in Figure \ref{fig:critical_examples} some images of the DeepFashion2 dataset \cite{deepfashion2} are shown where it is hard to determine if the picture shows a skirt or shorts, even for a human observer.
	We compared the prediction scores and their expressiveness of the SAE and an equivalent traditional classifier for skirts and shorts.
	We normalized the prediction scores provided by the SVM by scaling the scores between the class centers into the interval $[0...1]$.
	Fig. \ref{fig:confidence_comparison} shows the relation between the prediction scores and the actual precision.
	The noisy graph of the traditional classifier shows that the prediction score provides only a rough evidence about the class membership probability.
	For example the real precision of $0.4$ can be reflected by a prediction score between $0.25$ and $0.65$. 
	In contrast the stable and monotonous relation when using the SAE shows that its prediction scores reflect the uncertainty much better. 
	That means the confidence given by the SAE is much more reliable and expressive.
	In contrast softmax activations in combination with cross entropy loss let traditional classifiers tend to predict scores that are either close to $1$ or $0$ as seen in Figure \ref{fig:score_frequency}.
	Confidences between these extremes are mostly noisy with a low informative value.
	Structuring Autoencoders do not suffer from this drawback since they naturally achieve a smooth separation of the classes and make use of the reconstruction loss given by both the labeled and unlabeled samples.
	Regarding only classification tasks the reconstruction loss can also be interpreted as a regularization term for the structural loss function.
	
	\subsection{Guided Labeling}
	Since the SAE combined with an SVM provides a reliable decision confidence it can be used to efficiently discover important samples in the test set.
	After projecting into the latent space samples with a high uncertainty for a class do not show any exceptionally high SVM classification score compared to the rest of the classes.
	We identify these critical samples by calculating the scores for each class and compare the highest score to the second highest score.
	A small difference indicates a high uncertainty.
	The most important of these data points under this criterion can then be labeled manually and included in the training data.
	This guides the training process such that only a small amount of data needs to be labeled.
	To achieve a realistic setting we did not delete the points from the test set but instead define an unlabeled set of samples from the training set of the respective datasets.
	Note that misclassified data points are not detected by this method.
	However, our experiments show that the classification performance significantly improves on the unchanged test set which means formerly misclassified samples are now correctly classified.
	Fig.~\ref{fig:guided_sampling} shows the performance of a SAE combined with an SVM classifier initially trained with $600$ samples for $200$ epochs on MNIST.
	In epoch $200$ the $100$ most important data points from the unlabeled set are automatically detected and included in the training set.
	This results in a decrease of the classification error from approximately $4$\% to $3$\%.
	It is compared against a SAE trained with randomly sampled data to show that the better performance is a result of the intelligent choice of new samples and not of the increased number of samples.
	Additionally, we show that our methods outperforms a neural network of the same structure as the encoder part of the SAE which is solely trained for classification.
	Using the guided labeling approach the time and cost for manual annotations can be significantly reduced since only the most important samples (i.e. the samples with the highest uncertainty) need to be labeled manually.
	
	
	\subsection{Effect of MDS}
	\label{effect_mds}
	As stated earlier our modification to a standard autoencoder training only has a minor influence on the autoencoders reconstruction.
	This influence is regulated by the parameter $\gamma$ in Eq.~\ref{eqn:loss}, where $\gamma =0$ means that the structural loss is ignored during training, i.e. a traditional autoencoder is trained.
	Setting $\gamma = 1$ means only the structural loss is considered.
	Fig.~\ref{fig:gamma_plot} shows the reconstruction error and the classification error on the three datasets with different values for $\gamma$. 
	Assuming that a low reconstruction error and a low classification error is desired we can estimate the best values for $\gamma$ in Fig.~\ref{fig:gamma_plot} as $0.5$ for MNIST and $0.75$ for Fashion-MNIST.
	The best value for 3D HumanPose lies around $0.004$\footnote{This low weight can be explained by the numerical low reconstruction error as seen in Fig.~\ref{fig:gamma_plot}.}.
	The reconstruction error does not increase much when applying the structural loss.
	That means the reconstructions remain equally good for a wide range of values for $\gamma$.
	
	Having a closer look at the results in Fig.~\ref{fig:gamma_plot} for MNIST and Fashion-MNIST reveals a slight rise when $\gamma$ gets close to $1$ (i.e. the network is mostly optimized for classification).
	
	This underlines our claim that the SAE efficiently combines the natural structuring properties of traditional autoencoders with an additional structural information.
	
\begin{figure}[tp]
		\centering
		\includegraphics[width=0.42\textwidth]{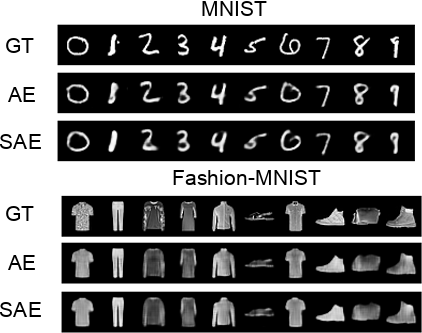} 
		\caption{Reconstructions of MNIST and Fashion-MNIST obtained by the SAE compared to ground truth and a standard autoencoder. The SAE produces a quality of output images that is comparable and in some cases subjectively better compared to the traditional autoencoder.}
		\label{fig:rec_mnist_cifar}
	\end{figure}

\begin{figure}
		\vspace*{-10pt}
		\centering
		\includegraphics[width=0.48\textwidth]{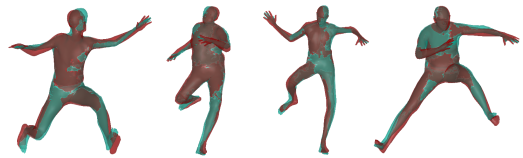} 
		\caption{Reconstructions (green) obtained by the SAE of the 3D body shapes compared to ground truth (red). Body shape and pose are reconstructed well. Only minor deviations can be seen in the extremities.}
		\label{fig:rec_smpl}
		\vspace*{-8pt}
	\end{figure}

	For subjective evaluation Fig.~\ref{fig:rec_mnist_cifar} shows some example reconstructions for MNIST and Fashion-MNIST while Fig.~\ref{fig:rec_smpl} shows examples for 3D HumanPose.
	The reconstructions of the SAE and the traditional autoencoder are nearly indistinguishable.
	
	\subsection{Class Transitions}
\begin{figure}[tp]
		\centering
		\includegraphics[width=0.42\textwidth]{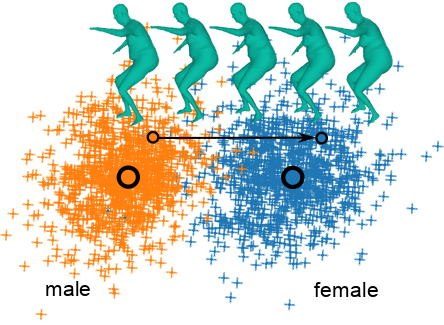} 
		\caption{Visualization of the body shape morphing in the latent space. The two classes male and female are well separated. When adding the directional vector defined by vector between the centers of the male and female clusters the male body shape clearly transitions to female while maintaining the body pose.}
		\label{fig:morph_vis}
	\end{figure}

	By exploiting the separated latent space it is possible to transition from one class to another.
	For visualization we use the 3D HumanPose dataset and the corresponding autoencoder trained to separate into male and female body shapes.
	The deformation vector is defined by the vector from the class center of the female class to the center of the male class or vice-versa.
	To morph between classes the scaled deformation vector is added to the latent variables.
	The morphed reconstruction is then obtained by applying the decoder to the changed latent variables.
	The step-wise morphing from male to female is visualized in Fig.~\ref{fig:morph_vis}.
	As can be seen there is a smooth transition between the classes.
	Interestingly the body pose does not change much while morphing. 
	That means the autoencoder learns to structure the latent space for the pose component by itself.
	Moreover, this structure seems to be similar for the male and female clusters in the latent space.
	This underlines our claim that the self structuring properties of traditional autoencoders can be efficiently combined with another given structure using the SAE.
	
	\section{Conclusion}
	We presented a method to improve traditional autoencoders such that they are able to structure the latent space according to given labels.
	Our SAE is able to separate different classes in the latent space even if this separation is not present in the data.
	By combining the traditional Multidimensional Scaling technique with novel autoencoder architectures the latent space is not only well structured but also preserves predefined distances between the different classes.
	We showed that a simple linear classifier on the latent variables outperforms comparable neural networks in classification tasks.
	In sparsely-supervised settings the SAE helps lowering the amount of required training data to reduce labeling cost and time.
	At the same time the prediction of unknown samples is more interpretable which, unlike standard classifiers, enables a reliable decision confidence.
	Based on this we developed a guided labeling approach by exploiting distances to class boundaries in the latent space which detects the unlabeled data points with the highest classification uncertainty.
	Additionally, an example for the combination of the self structuring properties of traditional autoencoders with the proposed MDS method is shown.
	Our proposed SAE could be used in the future to improve tasks like human pose estimation \cite{Wandt_2019_CVPR} and anomaly detection \cite{YinLia2018}. Furthermore, it may be combined with Markov Chain Neural Networks \cite{AwiRos2018a}.
	
	
	\newpage
	
	{\small
		\bibliographystyle{ieee}
		\bibliography{main}
	}

\end{document}